\documentclass[reprint, cha]{revtex4-1}

\usepackage{amssymb}
\usepackage[english]{babel}    
\usepackage[dvips]{graphics}
\usepackage{subfigure}
\usepackage[pdftex]{graphicx}
\usepackage{multirow}
\usepackage{rotating}
\usepackage{picinpar}
\usepackage{color}
\usepackage{lineno}
\usepackage{listings}

\begin{document}

\title{Fractal descriptors based on the probability dimension: a texture analysis and classification approach}

\author{Jo\~ao Batista Florindo}
 	     \email{florindo@ursa.ifsc.usp.br}
\affiliation{Scientific Computing Group, S\~ao Carlos Institute of Physics, University of S\~{a}o Paulo (USP),  cx 369 13560-970 S\~{a}o Carlos, S\~{a}o Paulo, Brazil - www.scg.ifsc.usp.br} 

\author{Odemir Martinez Bruno}
              \email{bruno@ifsc.usp.br}
\affiliation{Scientific Computing Group, S\~ao Carlos Institute of Physics, University of S\~{a}o Paulo (USP),  cx 369 13560-970 S\~{a}o Carlos, S\~{a}o Paulo, Brazil - www.scg.ifsc.usp.br}

\date{\today}

\begin{abstract}
In this work, we propose a novel technique for obtaining descriptors of gray-level texture images. The descriptors are provided by applying a multiscale transform to the fractal dimension of the image estimated through the probability (Voss) method. The effectiveness of the descriptors is verified in a classification task using benchmark over texture datasets. The results obtained demonstrate the efficiency of the proposed method as a tool for the description and discrimination of texture images.
\end{abstract}


\maketitle

\section{Introduction}

Fractals have played an important role in many areas with applications related to computer vision and pattern recognition \cite{SMS10,HWZ08,CDHLAB03,W08,TWZ07,LC10}, owing to their flexibility in representing structures usually found in nature. In such objects, we observe different levels of detail at different scales, which are described in a straightforward manner by fractals, rather than through classical Euclidean geometry.

Most fractal-based techniques are based on the concept of fractal dimension. Altough this concept was originally defined only for mathematical fractal objects, it contains some properties that make it a very interesting descriptor for any object in the real world. Indeed, fractal dimension measures how the complexity (level of detail) of an object varies with scale, an effective and flexible means of quantifying how much space an object ocupies, as well as important physical and visual properties of the object, such as luminance and roughness.

Fractal techniques include the use of Multifractals \cite{H01,LRAJ08,LGS00}, Multiscale Fractal Dimension \cite{MCSM02,CC00} and Fractal Descriptors \cite{BPFC08,BCB09,PPFVOB05,FBCB12}. Here we are focus on the last approach, which has demonstrated the best results in texture classification \cite{FB13}. The main idea of fractal descriptors theory is to provide descriptors of an object represented in a digital image from the relation among fractal dimensions taken at different observation scales, thus these values provide a valuable information on the complexity of the object, in the sense that they capture the degree of detail at each scale. In this way, fractal descriptors are capable of quantifying important physical characteristics of the structure, as the fractal dimension, but presenting a richer information than can be provided by a single number (fractal dimension).

Although fractal descriptors have demonstrated to be a promising technique, we observe that they are defined mostly on well-known methods to estimate the fractal dimension. Here, we propose fractal descriptors based on a less known definition of fractal dimension: the probability dimension. This is a statistical approach, which measures the distribution of pixel intensities along the image. In this way, such descriptors can express how the statistical arrangement of pixels in the image changes with the scale and how much such correlation approximates a fractal behavior. In this sense, our descriptor also measure the self-similarity and complexity of the image but upon a statistical viewpoint. This is a rich and not explored perspective, which is studied in depth in this work.

We use the whole power-law curve of the dimension and apply a time-scale transform to emphasize the multiscale aspect of the features. Finally, we test the proposed method over two well-known datasets, that is, Brodatz and Outex, comparing the results with another fractal descriptor approach showed in \cite{BCB09} and other conventional texture analysis methods. The results demonstrat that probability descriptors achieve a more precise classification than other classical techniques.

\section{Fractal Theory}
 
In recent years, fractal geometry concepts have been applied to the solution of a wide range of problems \cite{SMS10,HWZ08,CDHLAB03,W08,TWZ07,LC10}, mainly because conventional Euclidean geometry has severe limitations in providing accurate measures of real-world objects.

\subsection{Fractal Dimension}

The first definition of fractal dimension provided in \cite{M68}, is the Hausdorff dimension. In this definition, a fractal object is a set of points immersed in a topological space. Thus one can use results from Measure Theory to define a measure over this object. This is the Hausdorff measure expressed by
\begin{equation}
	H^{s}_{\delta}(X) = \inf{\sum_{i=1}^{\infty}{|U_{i}|^{s}\mbox{: {U$_{i}$ is a $\delta$-cover of X}}}},
\end{equation}
where $|X|$ denotes the diameter of $X$, that is, the maximum possible distance among any elements of $X$:
\begin{equation}
	|X| = \sup\{|x-y|:x,y \in X\}.
\end{equation}
Here, a countable collection of sets ${U_i}$, with $|U_i| \leq \delta$, is a $\delta$-cover of $X$ if $X \subset \cup_{i=1}^{\infty}U_i$.

Notice that $H$ also depends on a parameter $\delta$, which expresses the scale at which the measure is taken. We can eliminate such dependence by applying a limit over $\delta$, defining in this way the $s$-dimensional Hausdorff measure:
\begin{equation}
	H^s(X) = \lim_{\delta \rightarrow 0}H^s_{\delta}(X).
\end{equation}
The plot of $H^s(X)$ as a function of $s$ shows a similar behavior in any fractal object analyzed. The value of $H$ is $\infty$ for any $s < D$ and it is $0$ for any $s > D$, where $D$ always is a non-negative real value. $D$ is the Hausdorff fractal dimension of $X$. More formally,
\begin{equation}
	D(X) = \{s\} | \inf \left\{ s:H^{s}(X)=0 \right\} = \sup \left\{ H^{s}(X)=\infty \right\}.
\end{equation}

In most practical situations, the Hausdorff dimension is difficult or even impossible to calculate. Thus assuming that any fractal object is intrinsically self-similar, the literature shows a simplified version, also known as the similarity dimension or capacity dimension:
\begin{equation}\label{eq:FD}
	D = -\frac{\log(N)}{\log(r)},
\end{equation} 
where $N$ is the number of rules with linear length $r$ used to cover the object. 

In practice, the above expression may be generalized by considering $N$ to be any kind of self-similarity measure and $r$ to be any scale parameter. This generalization has given rise many methods for estimating fractal dimension, with widespread applications to the analysis of objects that are not real fractals (mathematically defined) but that present some degree of self-similarity in specific intervals. An example of such a method is the probability dimension, used in this work and described in the following section.

\subsection{Probability Dimension}

The probability dimension, also known as the information dimension, is derived from the information function. This function is defined for any situation in which we have an object occupying a physical space. We can divide this space into a grid of squares with side-length $\delta$ and compute the probability $p_m$ of $m$ points of the object pertaining to some square of the grid. The probability function is given by
\begin{equation}
	N_P(\delta) = \sum_{m=1}^{N}{\frac{1}{m}p_m(\delta)},
\end{equation}
where $N$ is the maximum possible number of points of the object inside a unique square. Here we use a generalization of teh above expression defined in the multifractal theory \cite{PV02}:
\begin{equation}\label{eq:probDF}
	N_P(\delta) = \sum_{m=1}^{N}{m^{\alpha}p_m(\delta)},
\end{equation}
where $\alpha$ is any real number.

The dimension itself is given as
\begin{equation}\label{eq:prob}
	D = -\lim_{\delta \rightarrow 0}\frac{\ln N_P}{\ln \delta}.
\end{equation}

When this dimension is estimated over a gray-level digital image $I:[M,N] \rightarrow \Re$, a common approach is to map it onto a three-dimensional surface $S$ as
\begin{equation}
	S = \{i,j,I(i,j)|(i,j) \in [1:M] \times [1:N]\}.
\end{equation}
In this case, we construct a three-dimensional grid of 3D cubes also with side-length $\delta$. The probability $p_m$ is therefore given by the number of grid cubes containing $m$ points on the surface divided by the maximum number of points inside a grid cube.
\begin{figure*}[htbp]
	\centering
	\begin{tabular}{cccc}
		\includegraphics[width=0.3\textwidth]{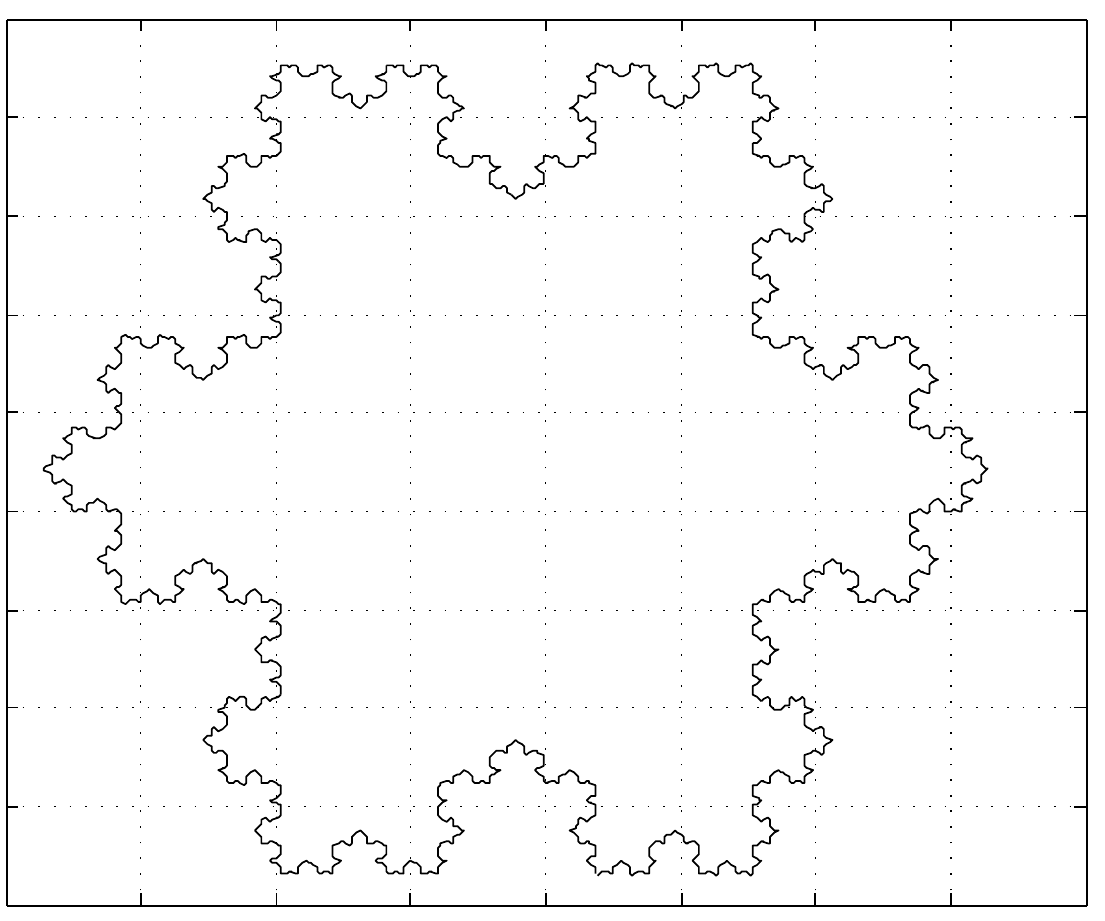} & \includegraphics[width=0.3\textwidth]{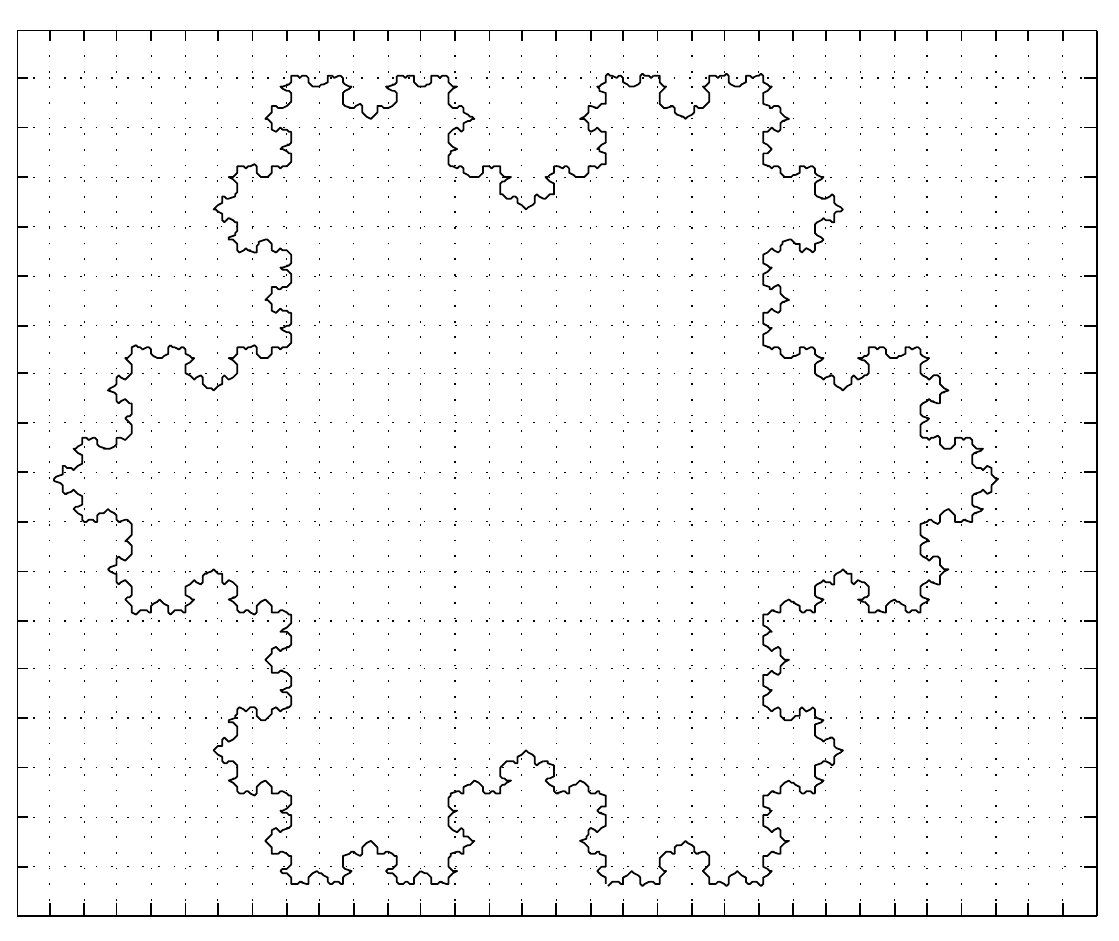} & \includegraphics[width=0.3\textwidth]{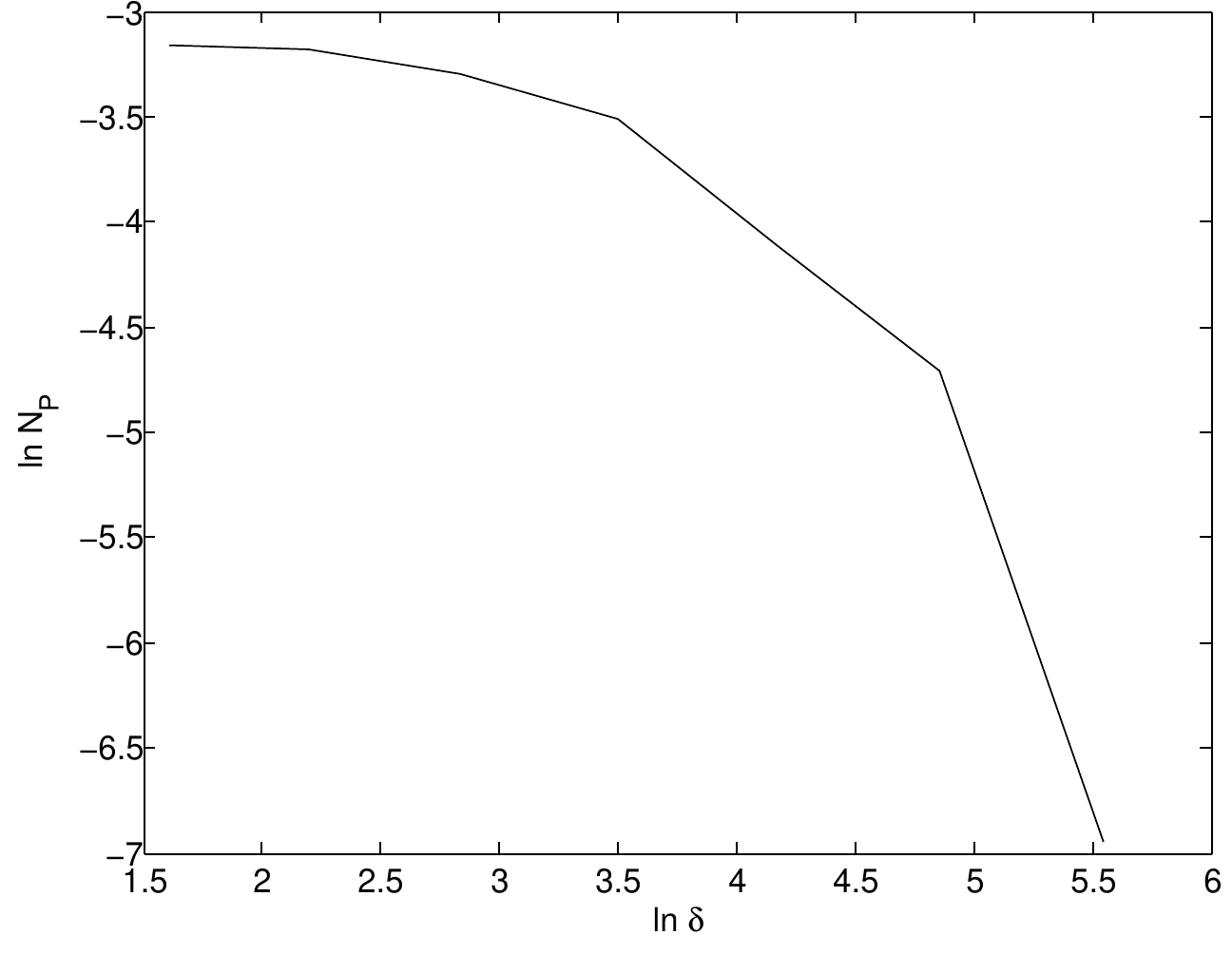}\\
		\includegraphics[width=0.3\textwidth]{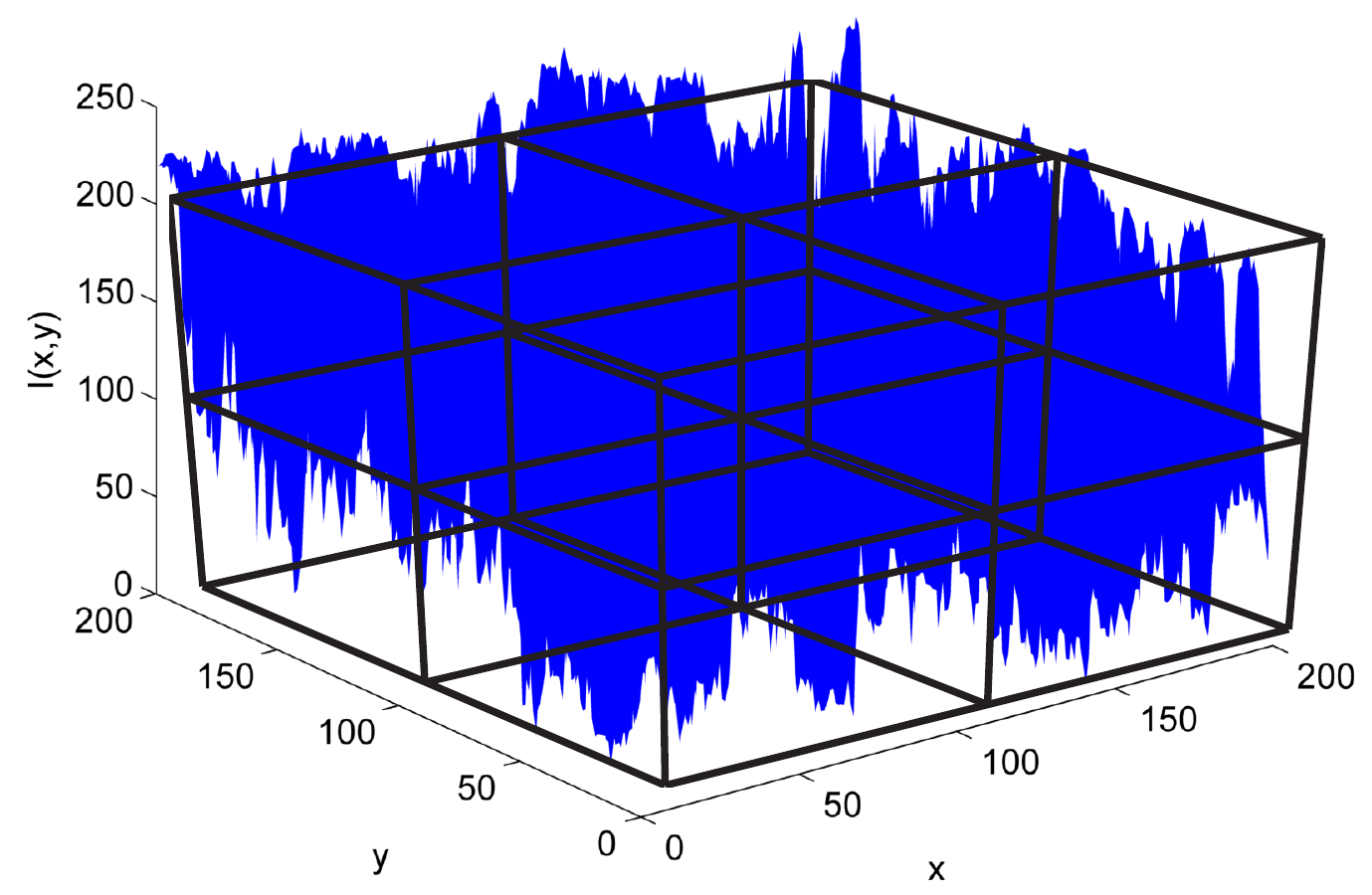} & \includegraphics[width=0.3\textwidth]{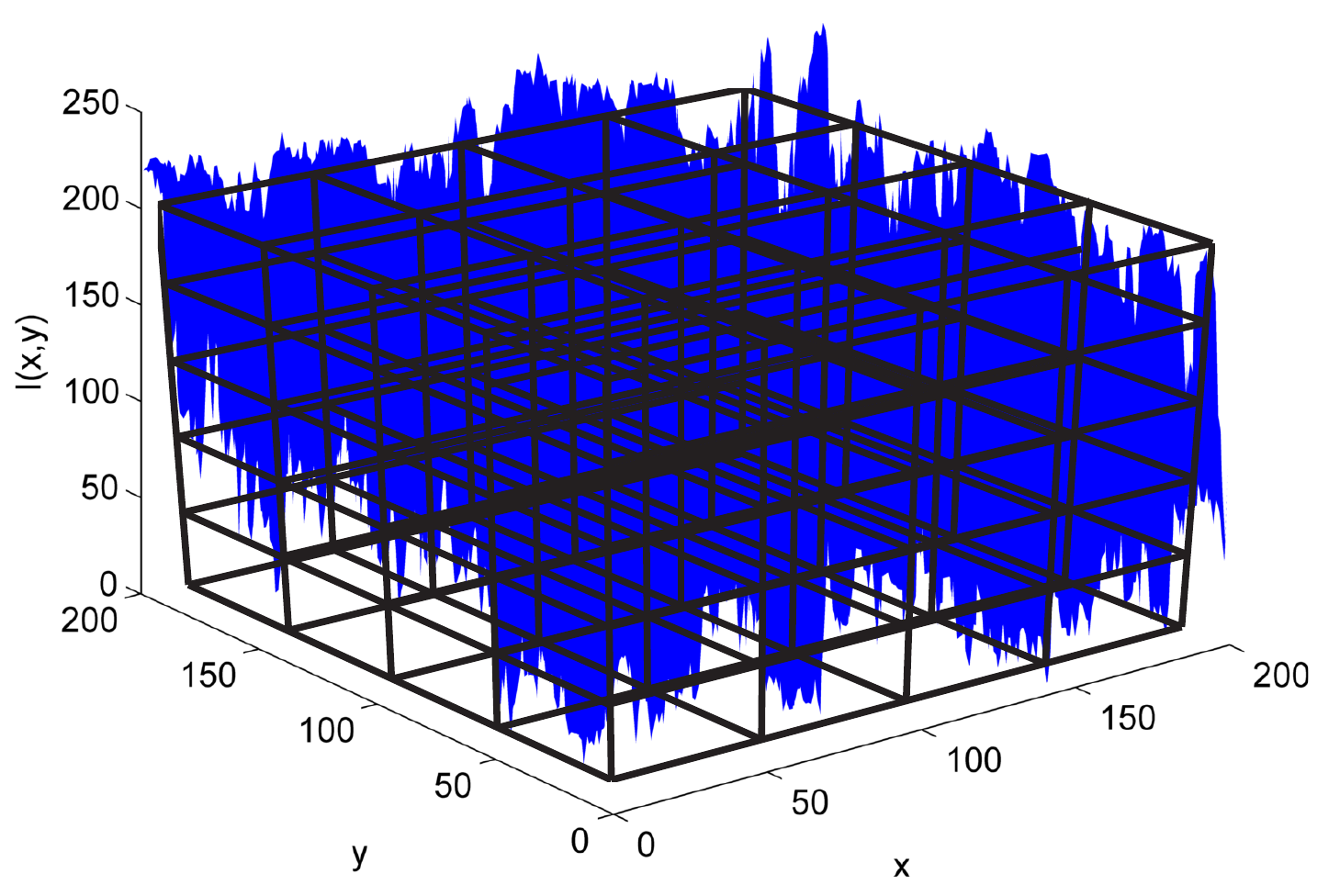} & \includegraphics[width=0.3\textwidth]{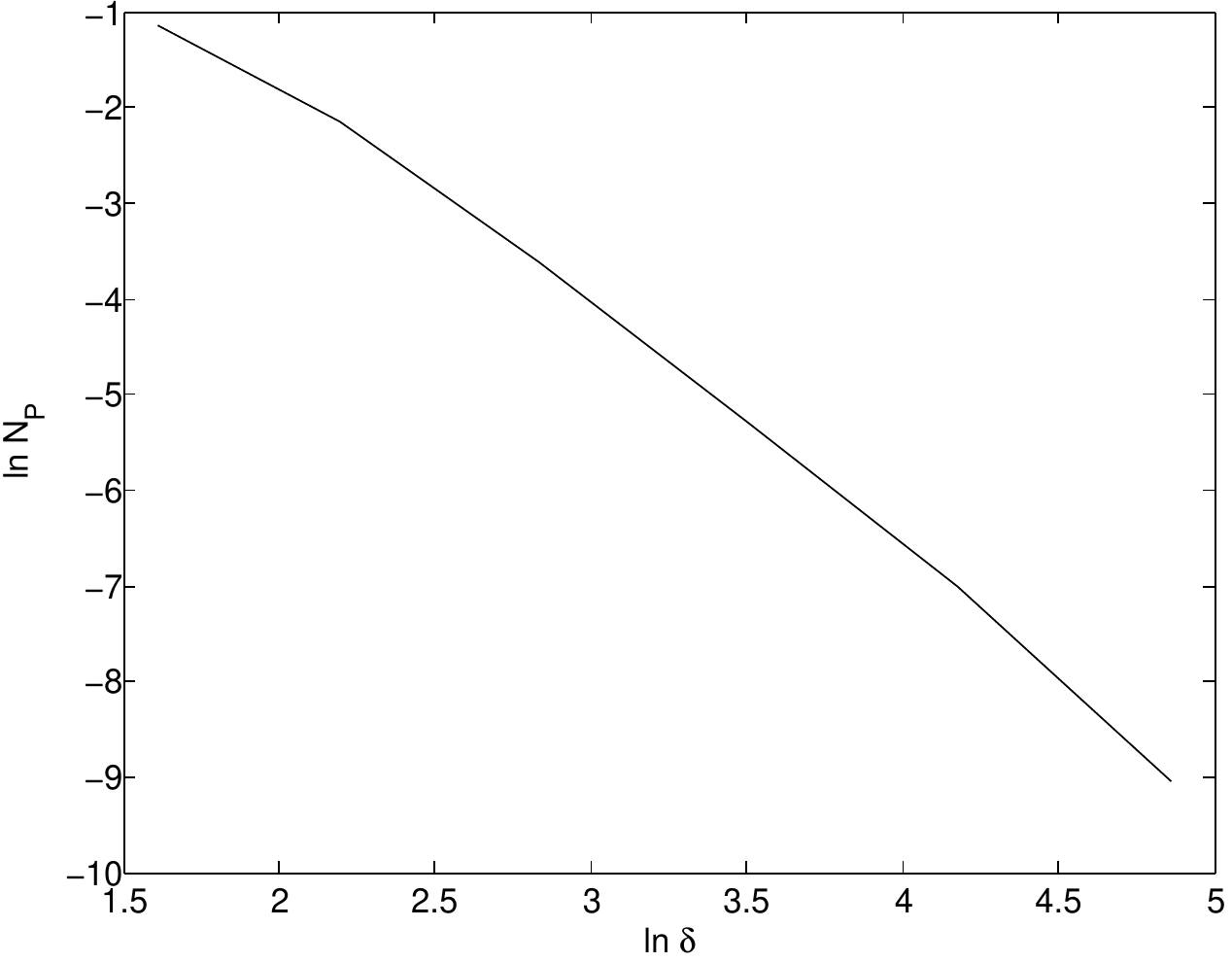}\\		
	\end{tabular}
	\caption{Two estimates of the probability dimension. Above, the 2D version used for shape analysis. Below, the 3D version used for gray-level images. In the 3D case, the original image is represented by a surface.}
	\label{fig:method}
\end{figure*}

\section{Fractal Descriptors}

Fractal descriptors are values extracted from the $\log-\log$ relationship common to most methods of estimating fractal dimension. Actually, any fractal dimension method derived from the concept of the Hausdorff dimension obeys a power-law relation, which may be expressed as
\begin{equation}\label{eq:FD2}
	D = -\frac{\log(\mathfrak{M})}{\log(\epsilon)},
\end{equation} 
where $\mathfrak{M}$ is a measure depending on the fractal dimension method and $\epsilon$ is the scale at which this measure is taken.

Therefore Fractal descriptors are provided from the function $u$:
\begin{equation}
	u:\log(\epsilon) \rightarrow \log(\mathfrak{M}).
\end{equation}
We call the independent variable $t$ to simplify the notation. Thus $t = \log \epsilon$ and our fractal descriptor function is denoted $u(t)$. For the probability dimension used in this work, we have
\begin{equation}
	u(t) = -\frac{\log(N_V(\delta))}{\log(\delta)}.
\end{equation}

\begin{figure*}[!htpb]
\centering
\includegraphics[width=0.8\textwidth]{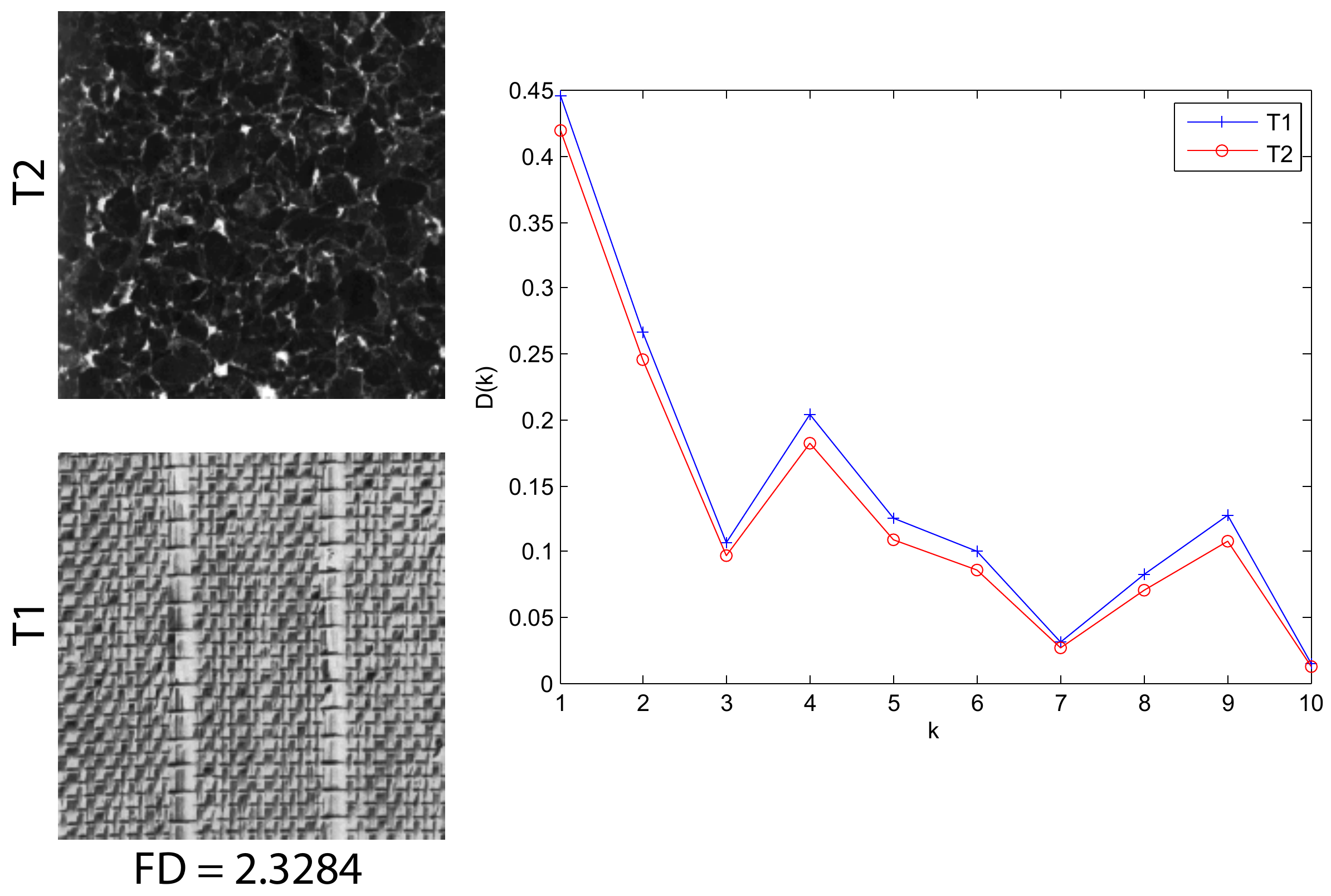}
\caption{Two textures with the same dimension (FD), but with different descriptors.}
\label{fig:DFText}
\end{figure*}

The values of $u(t)$ may be directly used as decriptors of the analyzed image or may be post-processed by some kind of operation aimed at emphasizing some specifical aspects of that function. Here, we apply a multiscale transform to $u(t)$ and obtain a bi-dimensional function $U(b,a)$, in which the variable $b$ is related to $t$ and $a$ is related to the scale at which the function is observed. A common means of obtaining $U$ is through a wavelet transform:
\begin{equation}
	U(b,a) = \frac{1}{a}\int_{\Re}{\psi(\frac{t-b}{a})u(t)dt},
\end{equation}
where $\psi$ is a wavelet basis function and $a$ is the scale parameter\cite{GM84}. Figure \ref{fig:DFText} shows an example where two textures with the same dimension, but visually distinct, provide different descriptors.

\section{Proposed Method}

This work proposes to obtain fractal descriptors from textures by using the probability fractal dimension, computing them from the curve $u(t):\log(N_P(\delta))$ in Eq. \ref{eq:prob}. Empirically, we obtained $0.2$ as the best value of $\alpha$ in the Equation \ref{eq:probDF}. Therefore we apply a multiscale transform to $u$.

The multiscale process employs a wavelet transform of $u(t)$, as described in the previous section:
\begin{equation}
	U(b,a) = \frac{1}{a}\int_{\Re}{\psi(\frac{t-b}{a})u(t)dt}.
\end{equation}

As the multiscale transform maps a one-dimensional signal onto a bi-dimensional function, it is a process that generates intrinsic redundancies. There are different approaches to elliminating such redundancies and keeping only the relevant information \cite{CC00}. Here, we adopt a simple method, fine-tuning smoothing, in which $U(b,a)$ is projected onto a specific value $a_0$ of the Gaussian parameter. We tested values of $a$ ranging between $0.1$ and $5$ and used the values that provided the best performance in the training experiments.

Finally, we selected a specific region from $U(b,a_0)$ to compose the descriptors. Empirically, we observed that the initial points in this curve provided better performance in our application. Then, we established a threshold $t$ after which all points in the convolution curve are disregarded and the values in the curve $U(i,a_0),1 \leq i \leq t$ are taken as the proposed descriptors.
\begin{figure}[!htpb]
\centering
\includegraphics[width=\columnwidth]{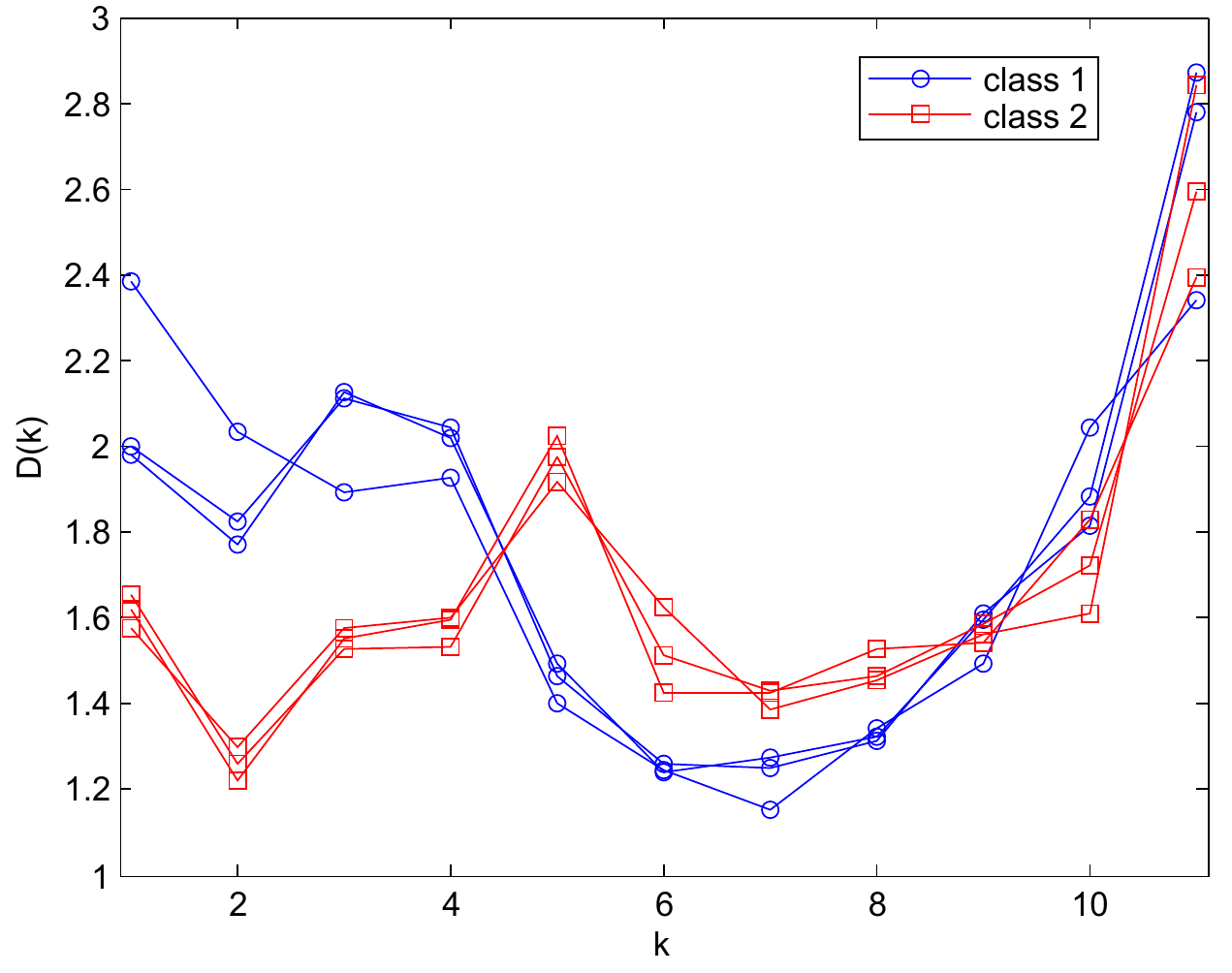}
\caption{Discrimination of texture image by the proposed descriptors. We have six images from two classes and the respective descriptors. Notice that the classes are substantially separated by the curves.}
\label{fig:classes}
\end{figure}

\section{Experiments}

In order to verify the efficiency of the proposed technique, we applied our probability descriptors to the classification of two benchmark datasets and compared our results to the performance of other well-known and state-of-the-art methods for texture analysis.

The first classification task used the Brodatz dataset, a classic set of natural gray-level textures photographed and assembled in an architecture book \cite{B66}. This dataset is composed by 111 classes with 10 textures in each class. Each image has a pixel dimension of 200$\times$200.

The second data set was Outex, a set of color textures extracted from natural scenes \cite{OMPVKH02}. Here, we used the first 20 classes, each one having 20 images with a 128$\times$128 pixel dimension, and converted them to gray-level images.

We compared our probability descriptors to six other techniques, namely, Local Binary Patterns (LBP) \cite{PHZA11}, Gabor-wavelets \cite{MM96}, Gray-Level Difference Method (GLDM) \cite{WDR76}, a multifractal approach described in \cite{PV02} and Bouligand-Minkowski fractal descriptors \cite{BCB09,FB13}.

Therefore we applied a Principal Component Analysis (PCA) \cite{DH00} over the data to elliminate or at least attenuate the correlation among the features. Finally we classified each descriptor by a K-fold process, with $K = 5$, using the Support Vector Machine (SVM) method \cite{V99} and compared the results.

\section{Results}

Table \ref{tab:brod} shows the correctness rate in the classification of the Brodatz dataset using the compared descriptors. The proposed method obtained the best result, outperforming the powerful Bouligand-Minkowski fractal descriptors and taking substantial advantage over other state-of-the art techniques such as Gabor and LBP. For this result we used $a = 0.1$ and a threshold $t = 8$. A particularly important aspect of our method with this data set is the reduced number of descriptors needed to provide a precise classification. This point is especially important in large databases, for which computational performance is more relevant. Furthermore, the small number of features avoids the curse of dimensionality, which impairs the reliability of the global result.
\begin{table*}[!htpb]
	\centering
	\caption{Correctness rate for Brodatz and Outex datasets.}
	\scriptsize
	\vspace{1em}
	Results for the Brodatz dataset
		\begin{tabular}{l|cc}
                 Method   & Correctness Rate (\%) & Number of descriptors\\
                 \hline
                  LBP & 82.5 $\pm$ 0.2 & 10\\
                  Gabor & 86.8 $\pm$ 0.1 & 8\\
                  GLDM & 79.6 $\pm$ 0.2 & 8\\
                  Multifractal & 54.9 $\pm$ 0.5 & 9\\
                  Bouligand-Minkowski & 90.8 $\pm$ 0.1 & 10\\
                  Proposed method & 93.0 $\pm$ 0.1 & 8\\
		\end{tabular}
	
		\vspace{1em}
		Results for the Outex dataset
		\scriptsize
		\begin{tabular}{c|cc}
                 Method   & Correctness Rate (\%) & Number of descriptors\\
                 \hline
                  LBP & 98.2 $\pm$ 0.0 & 7\\
                  Gabor & 98.5 $\pm$ 0.0 & 8\\
                  GLDM & 90.5 $\pm$ 0.1 & 9\\
                  Multifractal & 82.0 $\pm$ 0.2 & 10\\
                  Bouligand-Minkowski & 97.2 $\pm$ 0.0 & 10\\
                  Proposed method & 99.2 $\pm$ 0.0 & 7\\
		\end{tabular}
		
	\label{tab:brod}
\end{table*}

Table \ref{tab:brod} shows the results for the Outex textures. In this case, we obtained the best result by using $a = 0.1$ and $t = 7$. Again, the proposed approach provided the greatest success rate, despite the challenge of applying a gray-scale-based method to color analysis. In fact, Outex textures exhibit nuances which are better expressed in the color information, such as the changes in the lighting perspective and the images from different classes presenting similarities in the intensity distribution, though distiguished by color. Based on this result, our method demonstrates that although it does not use any color information, it is powerful also for color image analysis.

\begin{figure}
			 \centering
       \includegraphics[width=\columnwidth]{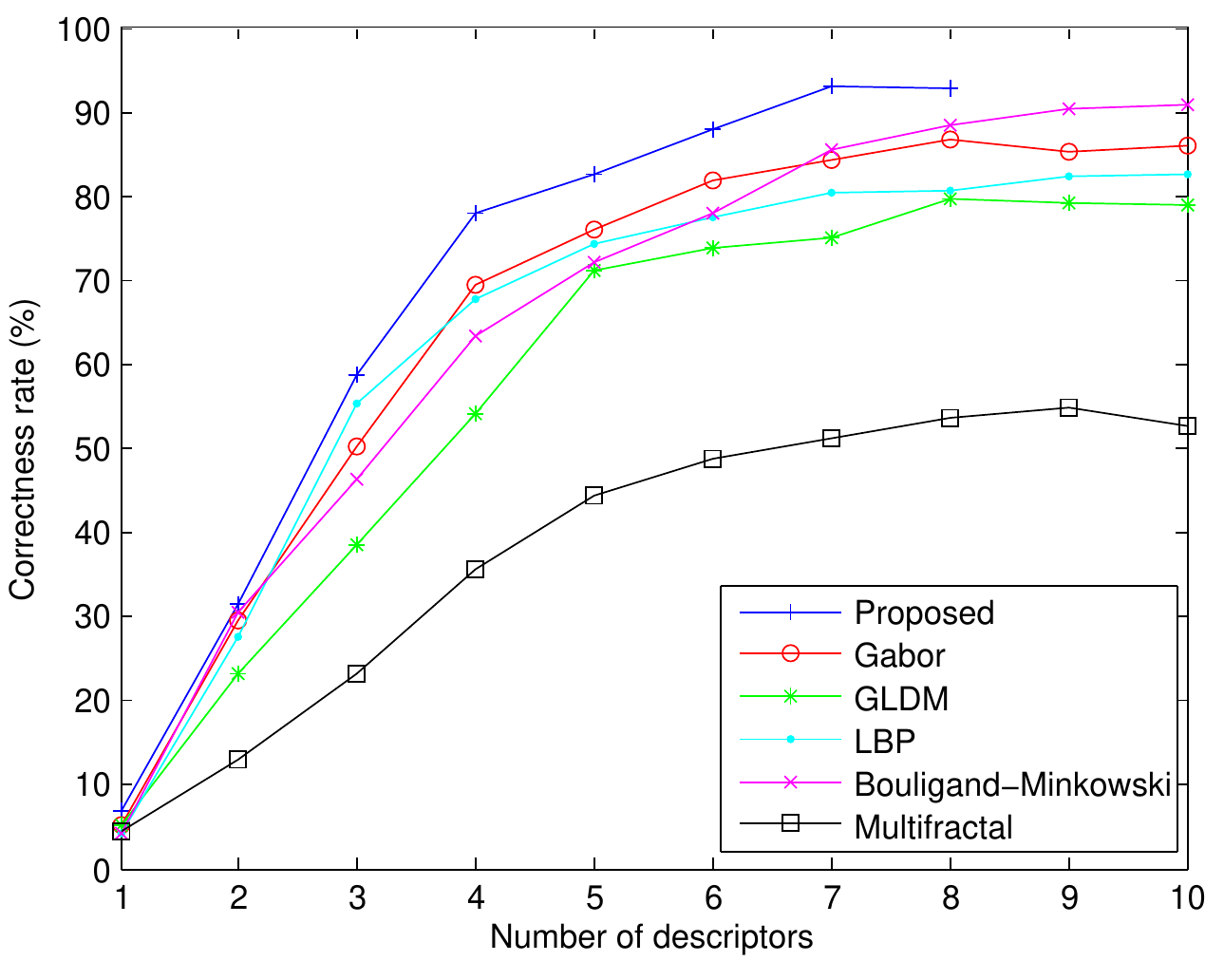}
            \includegraphics[width=\columnwidth]{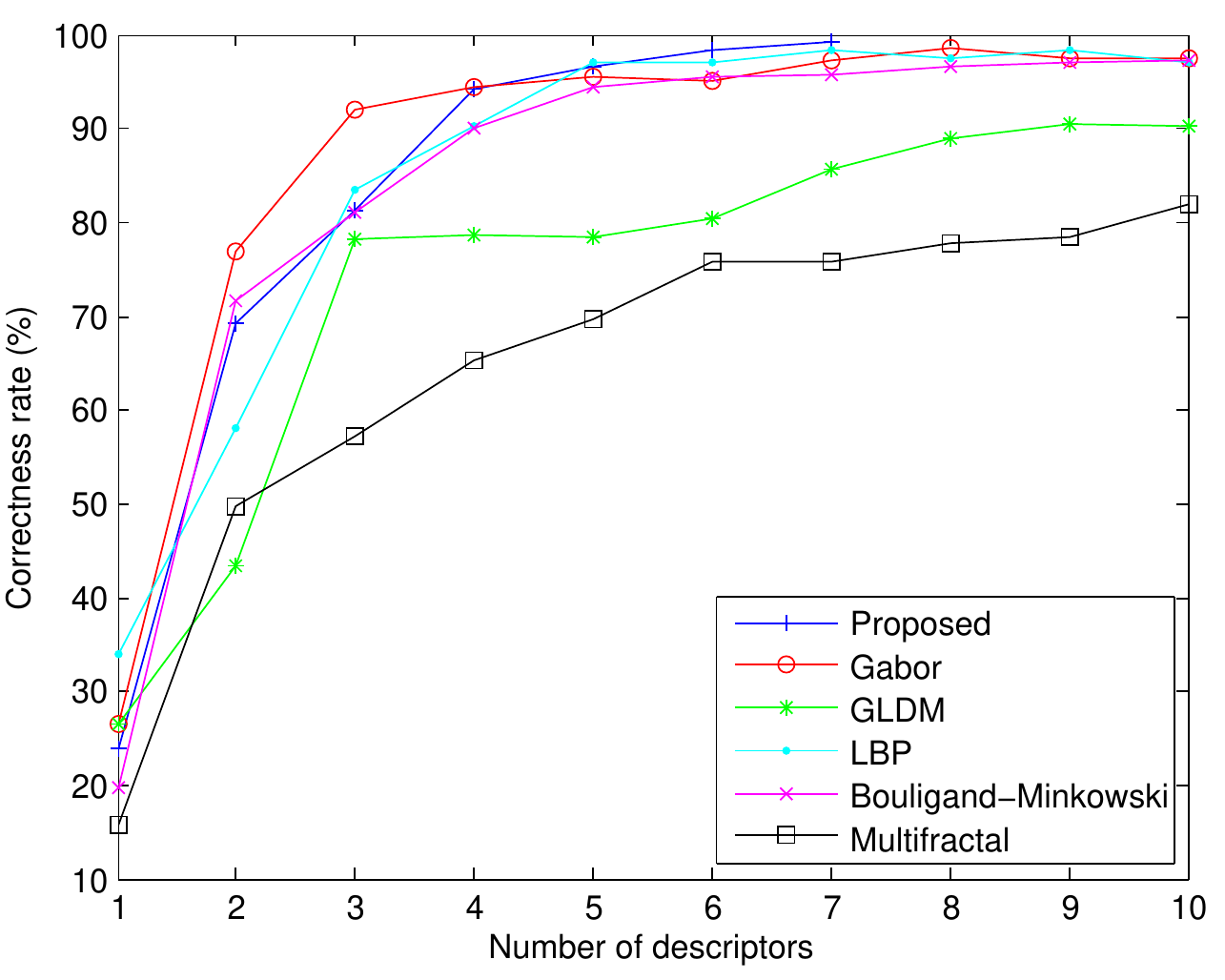}
       \caption{Success rate against the number of descriptors in each dataset. a) Brodatz. b) Outex.}
       \label{fig:acerto}                                  
\end{figure}

Figure \ref{fig:acerto} shows how the success rate varies according to the number of descriptors used in both datasets. The graphs show a well-known property of Karhunen-Lo\`{e}ve transform. The most expressive information is concentrated in the initial descriptors, so that the success curves show a quick growing and then tend to stabilize at a constant rate. The larger size and the native gray-scale format of Bradatz data leads to a clearer advantage of probability descriptors in that database. In Outex, the first descriptors, corresponding to the PCA components with higher variance, do not have as much significance for the classification purpose. However, the sum of all of them provide the best result. This is a specific property of fractal descriptors, as can be observed in the Bouligand-Minkowski descriptors for the Brodatz data as well. Fractal descriptors are tightly correlated among themselves, thus we do not have a large significance carried only in a few descriptors.

\begin{figure}
					 \centering
           \mbox{\subfigure[]{\includegraphics[width=0.5\columnwidth]{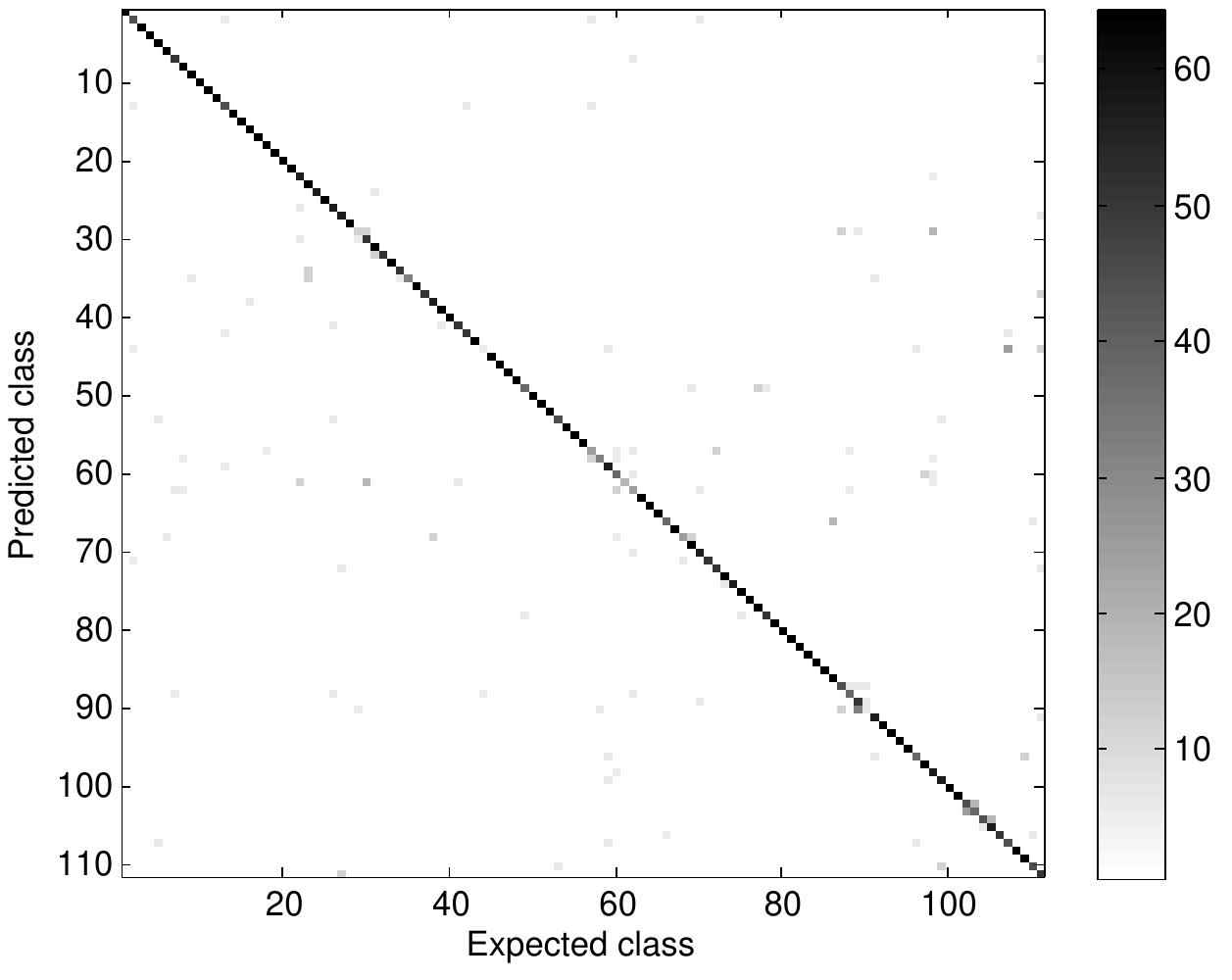}}
                 \subfigure[]{\includegraphics[width=0.5\columnwidth]{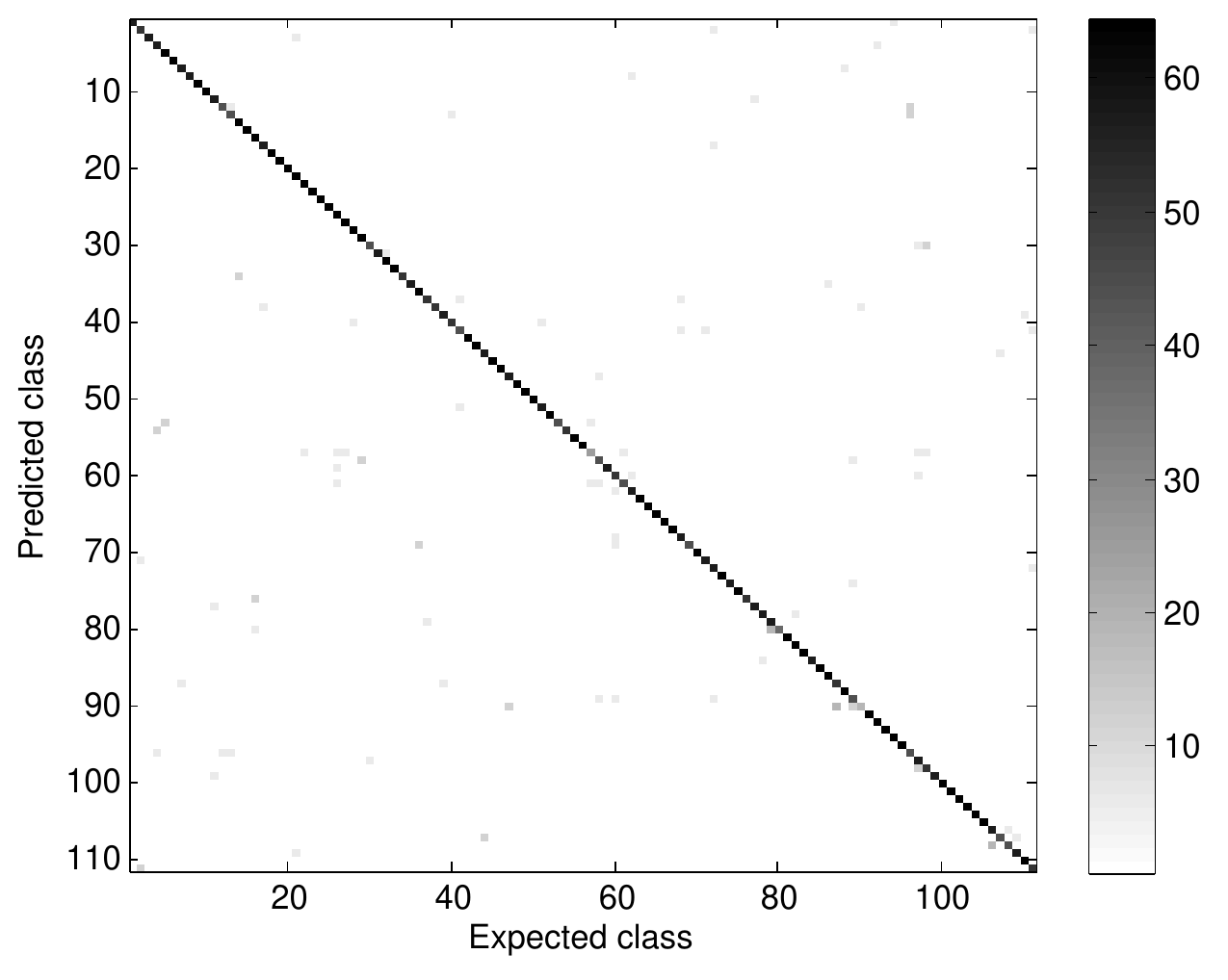}}}
           \mbox{\subfigure[]{\includegraphics[width=0.5\columnwidth]{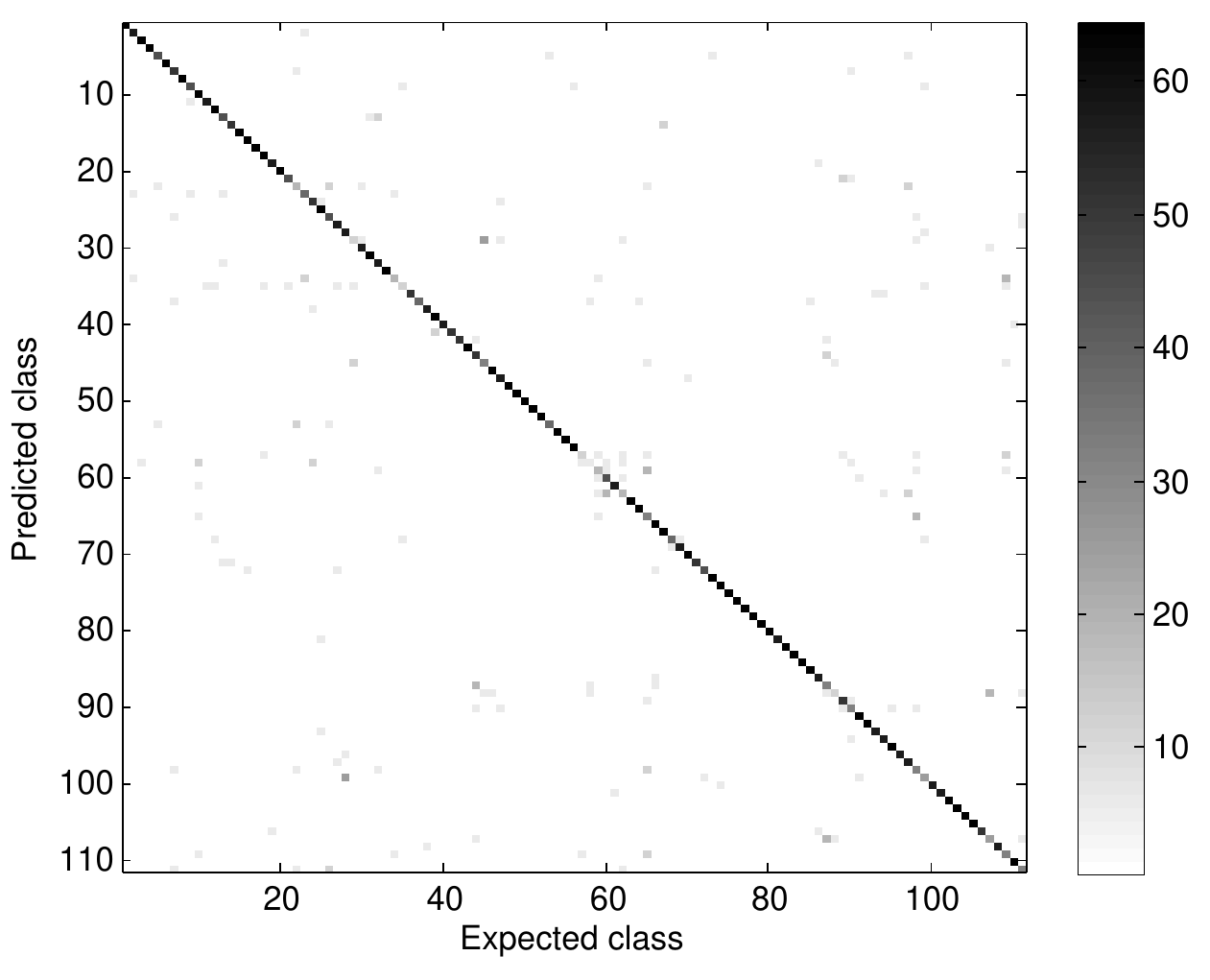}}
           			 \subfigure[]{\includegraphics[width=0.5\columnwidth]{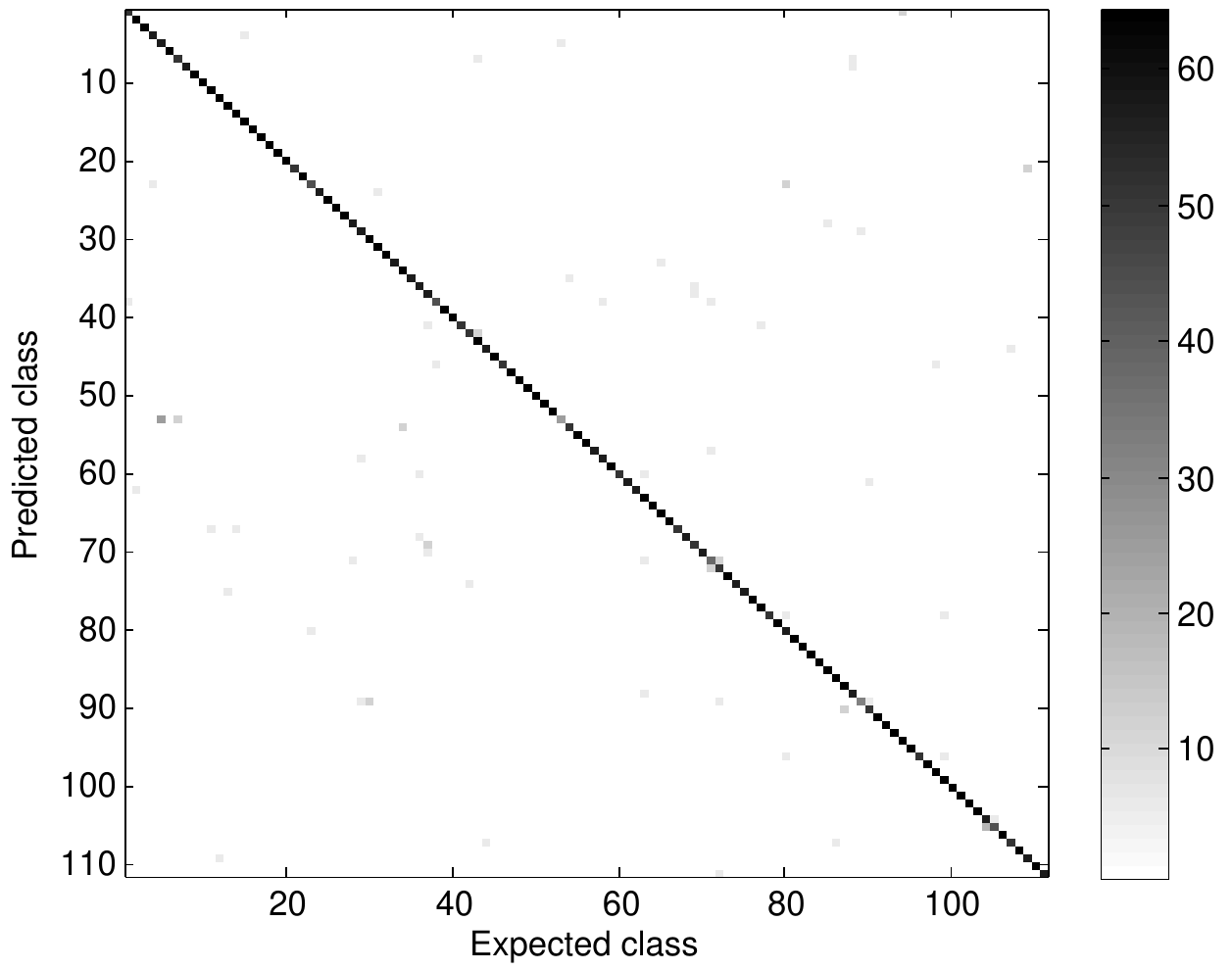}}}           			 
           \caption{Confusion matrices in Brodatz dataset. In such figures, we have the predicted classes in the rows and the expected ones in the columns. The number of images expected/predicted in each class is given by the gray-level at each point (lighter points correspond to large number of images). a) Gabor. b) Co-occurrence. c) LBP. d) Proposed method. }
           \label{fig:CMbrodatz}                                  
   \end{figure}

\begin{figure}
					 \centering
           \mbox{\subfigure[]{\includegraphics[width=0.5\columnwidth]{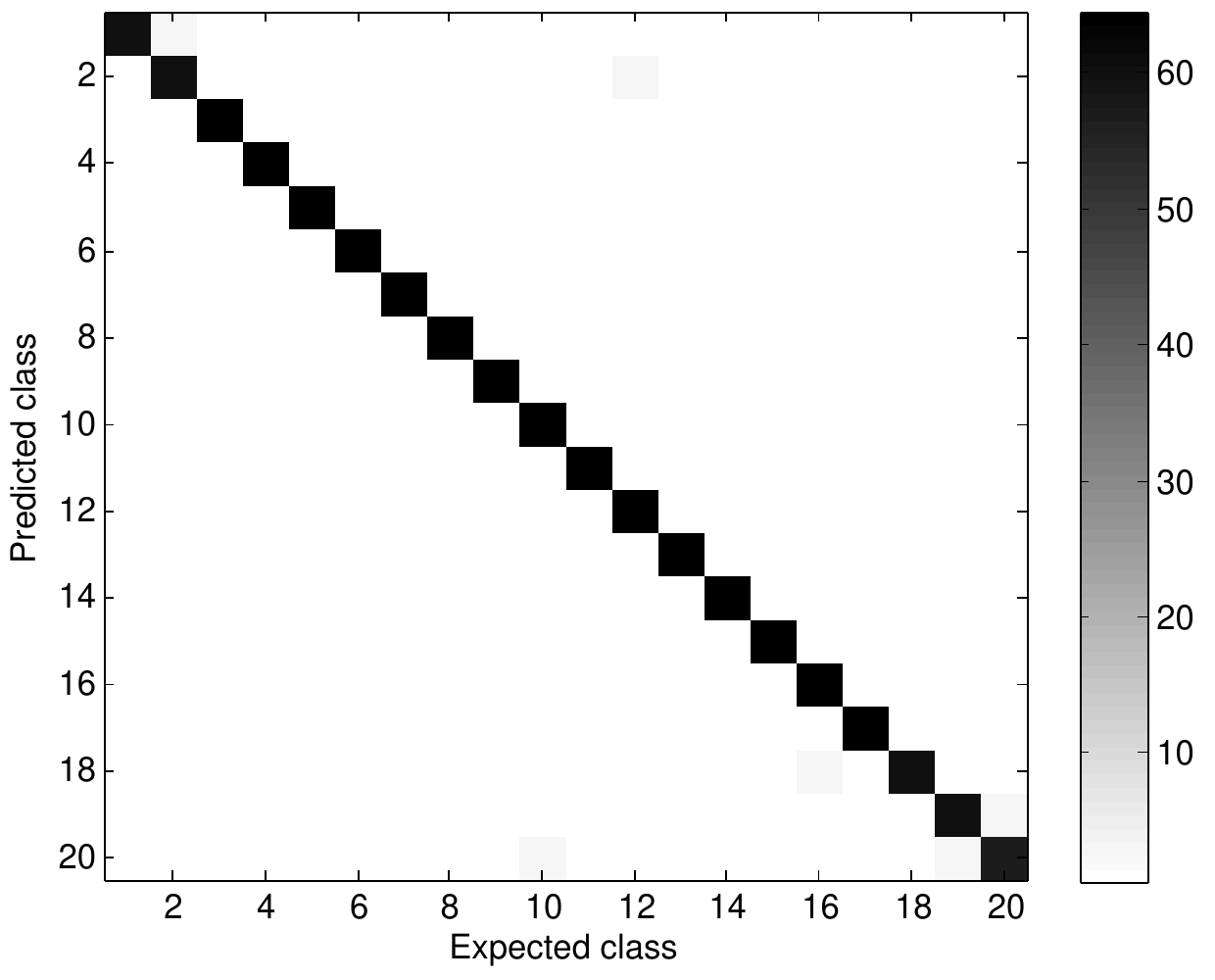}}
                 \subfigure[]{\includegraphics[width=0.5\columnwidth]{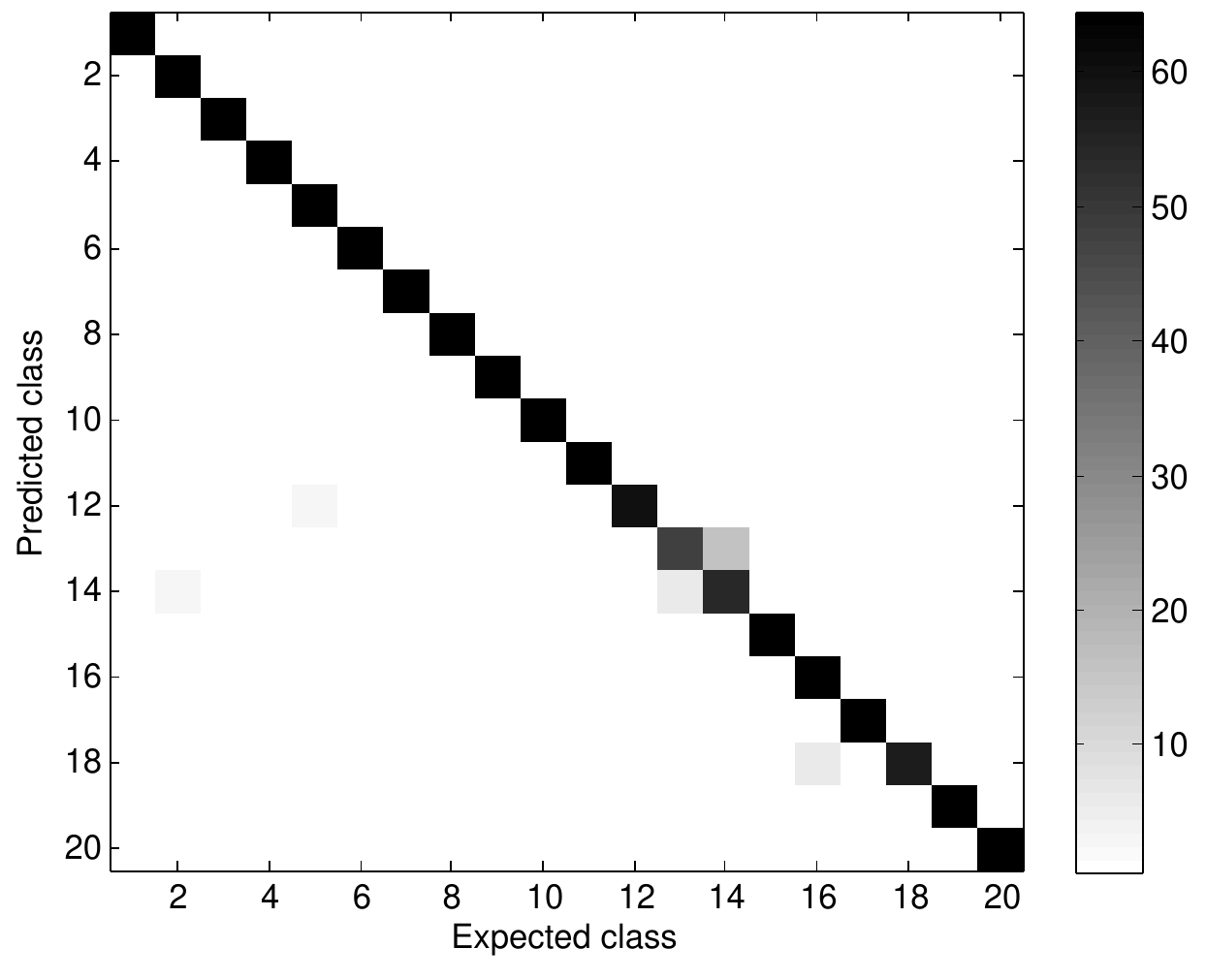}}}
           \mbox{\subfigure[]{\includegraphics[width=0.5\columnwidth]{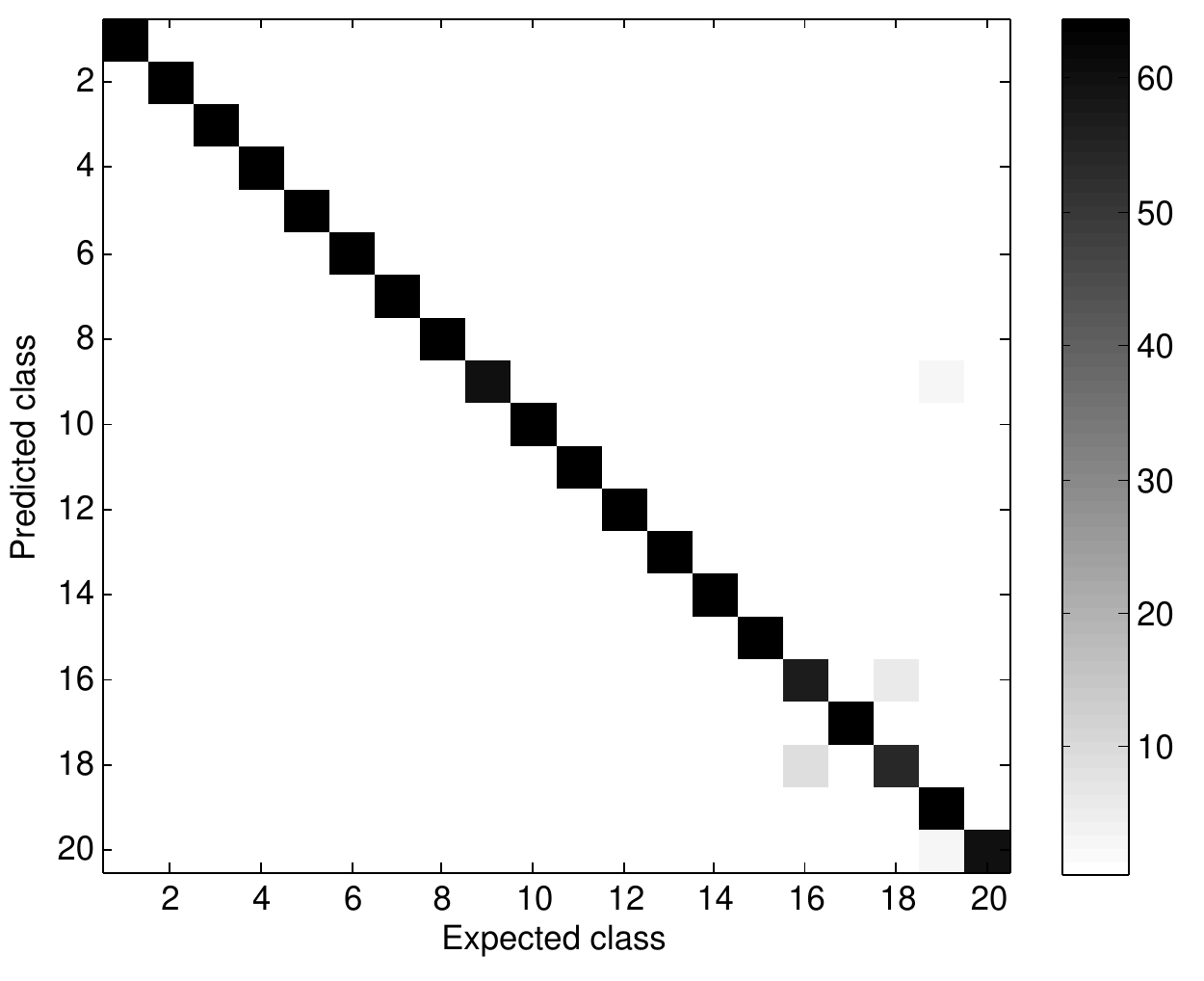}}
           			 \subfigure[]{\includegraphics[width=0.5\columnwidth]{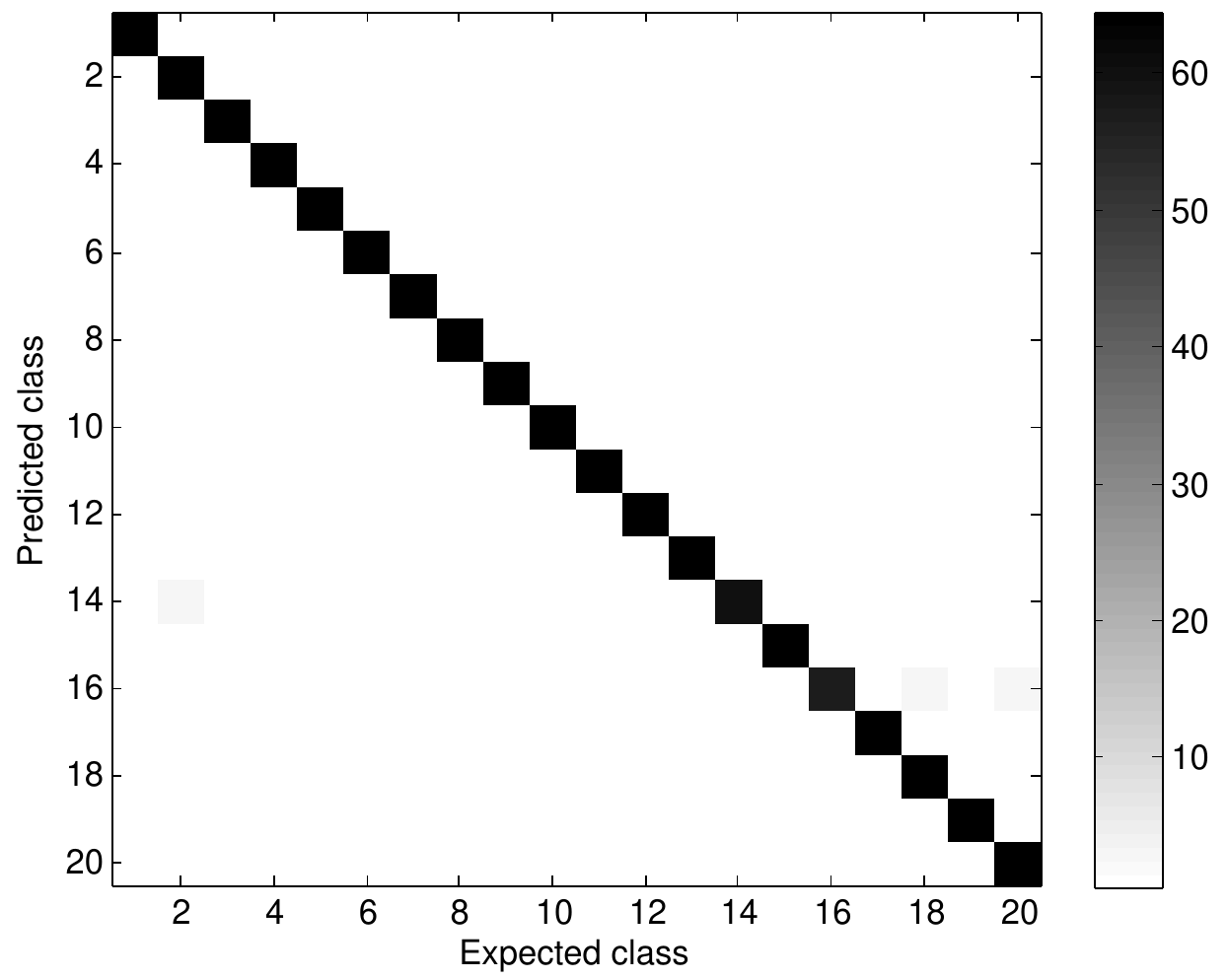}}}           			 
           \caption{Confusion matrices in Vistex dataset. a) Gabor. b) Co-occurrence. c) LBP. d) Proposed method. }
           \label{fig:CMvistex}                                  
   \end{figure}

Finally, Figures \ref{fig:CMbrodatz} and \ref{fig:CMvistex} show the confusion matrices of the methods with the best performances. In this kind of representation, a good descriptor must produce a matrix with a diagonal as lighter and continuous as possible and the minimum of dark points outside the diagonal. 

As can be seen, in Brodatz data, the probability descriptors clearly presented these characteristics, with almost no ``gap'' in the diagonal and with a few dark points outside. Both gaps and gray points indicate the confusion of the classifier, that is, elements classified incorrectly in some way. This confusion is caused mostly by the high similarity inter-class and low similarity intra-class. A precise descriptor, like the proposed, avoids such confusion by providing measures capable of faithfully representing the most complex structures.

In the case of Vistex, the diagonal gaps are not so clear, given the small number of classes. Thus the advantage of the proposed method can be seen in the reduced number of gray squares outside the diagonal. Such squares correspond to the confused classes. We observe that, particularly, the last classes have some discrimination difficulties. The elements of those classes are often assigned to other classes as depicted in the matrices. However, even in these cases, the proposed descriptors showed the expected robustness, assigning the elements correctly.
   
An overall analysis of the results demonstrates that the proposed method outperformed the compared ones in both datasets, using a small number of descriptors. Such results were expected from fractal theory given its wide applicability to the analysis of natural textures. Actually, fractal geometry presents a remarkable flexibility in the modeling of objects that cannot be well represented by Euclidean rules. The fractal dimension is a powerful metric for the complex patterns and spatial arrangements usually found in nature. Fractal descriptors provide a way of capturing multiscale variations and nuances that could not be measured by conventional methods. More specifically, the probability descriptors proposed here combine a statistical approach with fractal analysis, comprising a framework that supports a precise and reliable discrimination technique, as confirmed in the above results.     

\section{Conclusion}

We have proposed a novel method for extracting descriptors by applying a multiscale transform over the power-law relation of the fractal dimension estimated by the probability method.

We tested the efficiency of the proposed technique in the classification of two well-known benchmark texture datasets and compared its performance to that of other classical texture analysis methods. The results demonstrated that probability fractal descriptors are a powerful tool for modeling such textures. The proposed method achieved a high success rate in the classification of the benchmark data sets, using fewer than 10 descriptors in this task. These results demonstrate that the proposed method is capable of combining precision, low computational cost and robustness.

As a consequence, our method offers a reliable approach to solve a large class of problems involving the analysis of texture images.

\section*{Acknowledgments}

J. B. Florindo gratefully acknowledges the financial support of FAPESP Proc. 2012/19143-3.
O. M. Bruno gratefully acknowledges the financial support of CNPq (National Council for Scientific and Technological Development, Brazil) (Grant \#308449/2010-0 and \#473893/2010-0) and FAPESP (The State of S\~ao Paulo Research Foundation) (Grant \# 2011/01523-1).

\newpage


\end{document}